\pdfoutput=1
\documentclass[11pt]{article}
\usepackage[preprint]{acl}
\usepackage{longtable}
\usepackage{enumitem}
\usepackage{geometry}
\usepackage{graphicx}
\usepackage{adjustbox}
\usepackage{amssymb}

\usepackage{subcaption}
\usepackage{array,subcaption}
\usepackage{booktabs}
\usepackage{ragged2e}

\usepackage{multirow}
\usepackage{multicol}
\usepackage{times}
\usepackage[most]{tcolorbox}
\usepackage{latexsym}
\usepackage{arydshln}
\usepackage{hyperref}
\usepackage{pifont}
\usepackage{float}
\usepackage{bm}
\usepackage{microtype}
\usepackage{inconsolata}
\usepackage{xcolor}
\usepackage{xspace}
\usepackage{svg}

\usepackage{placeins}
\usepackage{afterpage}
\usepackage{amsfonts}
\usepackage{amsmath}

\usepackage[T1]{fontenc}
\usepackage[utf8]{inputenc}
\title{Harnessing Test-time Domain Adaptation for NLU tasks Involving Dialects of English}

\author{Duke Nguyen, Aditya Joshi, Flora Salim
\\
        University of New South Wales, Australia
\\
       \textbf{Correspondence:} \href{mailto:research@itsduke.me}{research@itsduke.me}
}

\begin{document}
\maketitle
\begin{abstract}
Test-time domain adaptation (TTDA) is an excellent method which helps generalize models across domains, tasks, and distributions without the use of labeled datasets. Thus, TTDA is very useful in natural language processing (NLP) in the dialectal setting, since oftentimes, models are trained on Standard American English (SAE), evaluated on Indian English (IndE), Singaporean English (SingE), or Nigerian English (NgE), of which distribution differs significantly from the former. This is especially useful since dialectal datasets are scarce. In this paper, we explore one of the most famous TTDA techniques, SHOT, in dialectal NLP. We finetune and evaluate SHOT on different combinations of dialectal GLUE. \textbf{Our findings show that SHOT is a viable technique when labeled datasets are unavailable}. We also theoretically propose the concept of \textbf{dialectal gap} and show that it has a positive correlation with the effectiveness of SHOT. We also find that in many cases, finetuning on SAE yields higher performance than finetuning on dialectal data.
\end{abstract}


\section{Introduction}
Human languages exhibit natural variations with respect to various dimensions \cite{Zampieri_Nakov_Scherrer_2020}. One such important dimensions is diatopic, also commonly referred to as `dialects' (which we use interchangeably). Examples of diatopic varieties include: "I \textit{might could} help you with that`` (Southern US, Australian, and New Zealand English), "Inside tent \textit{can not see leh!}`` (Singaporean English) \cite{acmcsuraditya}.




These natural linguistic variations pose significant challenges to NLP models. Specifically, existing works have shown serious gaps with Large Language Models when processing Standard American English versus other forms of English. These gaps have been demonstrated, in the case of bias in hate speech datasets \citet{sap-etal-2019-risk}, dependency parsing \cite{blodgett-etal-2018-twitter}, etc. In addition, the most extensive English dialectal benchmark, DialectBENCH \cite{faisal-etal-2024-dialectbench} only covers a quarter of documented dialects of English (19/77) \cite{ewave}. Dialects are constantly changing and evolving, and it is infeasible to build a dataset for every single dialect.



Recent trends in computer vision (CV), however, show a promising new field, test-time domain adaptation (TTDA), which can overcome these challenges. TTDA can potentially adapt an LLM trained on one dialect to another unseen dialect during inference. In this paper, we study the use of one such TTDA method, SHOT~\cite{shot}, in dialectal adaptation. SHOT is a foundational TTDA technique which uses k-means clustering to align the embedding of a classifier from the source domain to the target domain. We are the first to study dialectal adaptation using a TTDA technique. We perform our evaluations across four dialects: SAE, IndE, NgE, and SingE. We generate the respective dialectal version of GLUE in each of the above English dialects using Multi-VALUE~\citep{ziems2023multi}, a robust English dialectal translation system. We compare SHOT performance against in-dialectal finetuning and cross-dialectal finetuning. Our results indicate that: (1) the wider the dialectal gap between the source dialect (the dialect the model is finetuned on) and the target dialect (the dialect the model is evaluated on), the more effective SHOT is; (2) SHOT consistently improves the performance of models finetuned on SAE datasets and commonly outperforms models in-dialectal finetuned models; (3) cross-dialectal finetuning sometimes outcompetes in-dialectal finetuning.

\section{Related Work}\label{sec:related}


\begin{figure*}
    \centering
    \includegraphics[width=1\linewidth]{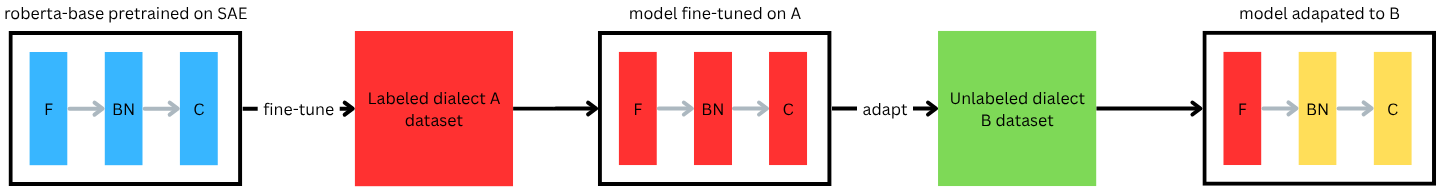}
    \caption{Dialectal SHOT pipeline. Blue indicates pretrained weights on SAE of roberta-base. Red indicates dataset or weights associated with source dialect A. Green indicates dataset or weights associated with target dialect B. Yellow indicates weights associated with adaptation from dialect A to B.}
    \label{fig:diagram}
\end{figure*}

There are limited English dialectal benchmark datasets. DialectBench~\cite{faisal-etal-2024-dialectbench} is large-scale NLP dialectal benchmark covering 10 text-level tasks and 281 varieties across languages. However, DialectBENCH is quite limited for the English language, being extensive only for question answering (SDQA). There are no other English dialectal benchmarks \cite{acmcsuraditya}. Most English dialectal research has been done with Multi-VALUE~\cite{ziems2023multi}, a dialectal translation system which applies documented syntactic structure of dialects as studied by \citet{ewave}. Although Multi-VALUE is a synthectic approach to create dialectal data, it is currently the state-of-the-art and used widely by many English dialectal papers \cite{held2023tada, liu-etal-2023-dada, srirag2024predicting, srirag-etal-2025-evaluating}. As a result, we use Multi-VALUE in our evaluation pipeline.

To overcome the dialectal gap, \citet{dacon-etal-2022-evaluating} apply adversarial learning to ensure consistence label prediction between the SAE and AAE inputs. \citet{tan-etal-2020-mind} propose a novel tokenization method which injects grammatical dialectical information as special symbols whilst reducing inflected words to their base form, improving the performance of AAE on SQUAD and MNLI. \citet{held2023tada} use an architecture of dialectal adapters and task adapters via a contrastive loss and morphosyntactic loss. \citet{xiao-etal-2023-task} use task-agnostic low-rank adapters to generalize an LLM to unseen dialects by leveraging expert linguistic knowledge via typological feature vectors. \citet{srirag2024predicting} similarly propose a task-agnostic low-rank adapter approach to the decoder model usecase. However, both rely on existing dialectal datasets.

\begin{table*}[ht]
    \centering
    
\begin{subtable}[t]{0.475\textwidth}
\centering
\subcaption{Evaluation on IndE Dialect}
\resizebox{\textwidth}{!}{%
\begin{tabular}{@{}llll@{}}

\toprule

S & CoLA                 & SST-2                 & RTE                   \\ \midrule

SAE           & \textbf{15.55} / \textbf{18.28} (2.73) & 87.52 / \textbf{87.89} (0.37)  & 58.57 / \textbf{59.76} (1.19)  \\

IndE        & 12.93                & \textbf{89.13}                 & 54.98                 \\

NgE      & 11.57 / 15.42 (3.85) & 86.02 / 87.02 (1.00)  & \textbf{58.96} / 55.78 (-3.18) \\

SingE   & 13.03 / 15.76 (2.73) & 87.52 / 86.64 (-0.88) & 54.98 / 54.98 (0)     \\

$\mu$       & 13.27 / 16.49 (3.10) & 87.55 / 87.18 (0.16)  & 56.87 / 56.84 (-0.66) \\ \bottomrule

\end{tabular}%
}
\end{subtable}%
\hfill
\begin{subtable}[t]{0.475\textwidth}
\centering
\subcaption{Evaluation on NgE Dialect}
\resizebox{\textwidth}{!}{%
\begin{tabular}{@{}llll@{}}

\toprule

S & CoLA                & SST-2                & RTE                   \\ \midrule

SAE                               & \textbf{19.51} / \textbf{21.53} (2.02)     & 88.01 / 88.89 (0.88)      & 58.96 / 60.16 (1.20)    \\

IndE                            & 10.41 / 13.57 (3.16)     & 88.64 / \textbf{90.14} (1.50)      & 54.18 / 52.99 (-1.19)   \\

NgE                          & 18.01                    & \textbf{89.63}                     & 58.96                   \\

SingE                       & 13.73 / 12.47 (-1.26)    & 87.52 / 88.26 (0.74)      & \textbf{60.16} / \textbf{61.75} (1.59)    \\

$\mu$                           & 15.42 / 15.86 (1.31)     & 88.45 / 89.10 (1.04)      & 58.07 / 58.30 (0.53)    \\ \bottomrule

\end{tabular}%
}
\end{subtable}%


\vspace{2ex} 

\begin{subtable}[t]{0.475\textwidth}
\centering
\subcaption{Evaluation on SingE Dialect}
\resizebox{\textwidth}{!}{%
\begin{tabular}{@{}llll@{}}

\toprule

S & CoLA                & SST-2                & RTE                   \\ \midrule

SAE           & 5.46 / \textbf{8.92} (3.47)  & 86.39 / 86.64 (0.25) & 58.17 / \textbf{58.96} (0.79)  \\

IndE        & 2.16 / 4.09 (1.93)  & 85.64 / 87.77 (2.13) & 53.78 / 54.98 (1.20)  \\

NgE      & 6.44 / 5.98 (-0.47) & 85.39 / 85.77 (0.38) & 57.77 / 56.57 (-1.20) \\

SingE   & \textbf{13.03}               & \textbf{88.38}                & \textbf{60.16}                 \\

$\mu$       & 6.77 / 6.33 (1.64)  & 86.45 / \textbf{86.73} (0.92) & 57.47 / 56.84 (0.26)  \\ \bottomrule

\end{tabular}%
}
\end{subtable}%
\hfill
\begin{subtable}[t]{0.475\textwidth}
\centering
\subcaption{Average Evaluation}
\resizebox{\textwidth}{!}{%
\begin{tabular}{@{}llll@{}}

\toprule

S & CoLA                 & SST-2                 & RTE                   \\ \midrule

SAE           & \textbf{13.51} / \textbf{16.24} (2.74) & 87.31 / 87.81 (0.50)  & \textbf{58.57} / \textbf{59.63} (1.06)  \\

IndE        & 8.50 / 8.83 (2.54)   & 87.80 / \textbf{88.96} (1.82)  & 54.31 / 53.99 (0)     \\

NgE      & 12.01 / 10.70 (1.69) & 87.01 / 86.40 (0.69)  & 58.56 / 56.18 (-2.19) \\

SingE   & 13.26 / 14.12 (0.74) & \textbf{87.81} / 87.45 (-0.07) & 58.43 / 58.37 (0.80)  \\

$\mu$       & 11.82 / 12.89 (2.02) & 87.48 / 87.67 (0.71)  & 57.47 / 57.04 (-0.08) \\ \bottomrule

\end{tabular}%
}
\end{subtable}%

\caption{Matrix comparison of models fine-tuned and applied with TTA on different dialects then evaluated on other dialects. $S$ stands for source dialect. For each cell, the fine-tuning value is on the left side, followed by the TTA value, with the TTA utility in brackets (). IndE stands for Indian English, NgE stands for Nigerian English, and SingE stands for Singaporean English. $\mu$ being the average performance. CoLA SAE on SAE performance is 49.44. SST-2 SAE on SAE performance is 91.97. RTE SAE on SAE performance is 58.12. Top values are boldened.}
\label{table:main-result}

\end{table*}

\section{Method}
Our model architecture is adapted from \citet{shot} for NLP\footnote{Our SHOT implementation is based on \citet{shot} at \url{https://github.com/tim-learn/SHOT}}, taking inspiration from \citet{zhang-etal-2021-matching}. SHOT was the first method proposed to address the problem of TTDA \cite{shot}. It is composed of three parts: (1) source model generation, (2) source hypothesis transfer with information maximization, and (3) source hypothesis transfer augmented with self-supervised pseudo-labeling. They are described as follows:

\begin{enumerate}[leftmargin=*, label=(\arabic*)]
    \item We train the source model with label smoothing using a training dataset from the source distribution. The source model is composed of an embedding network $F$, followed by a bottleneck network $BN$, then a classifier layer $C$. In the NLP setting, $F$ would be a transformer-based Encoder model, in our experiment, this is $\texttt{roberta-base}$. The bottleneck network $BN$ reduces the dimension of the output of $F$. We choose $BN$ to be a 2-layer bidirectional GRU~\cite{cho2014learning} and $C$ being a FFN, following similar works done by \cite{zhang-etal-2021-matching}. This is as shown on Figure \ref{fig:diagram} in the first two boxes.
    \item We follow the same settings as \citet{zhang-etal-2021-matching} and freeze $F$ to learn $BN$ and $C$ whilst minimizing the Information Maximization Loss which composes of the following two losses in Equation \ref{eq:loss} where $f_t(x) = C_t(BN_t(F_t(x)))$ being the $K$-dimensional output of each target sample, $1_K$ being the $K$-dimensional vector with all ones, and $p = E_{x_t \in X_t}[d(f_t^{(k)}(x_t))]$ is the mean output embedding of the whole target dialect. On an $L$ number of epochs, $C$ generates a pseudo-label, which is used to trained the model using the Information Maximization Loss. The pseudo-label is updated every iteration. This is as shown on Figure \ref{fig:diagram} in the last three boxes.
    \item Additionally, a weighted k-means clustering scheme is applied on the embedding $BN$ and $C$ to improve the quality of the pseudo-labels generated.
\end{enumerate}

{\footnotesize
\begin{equation}\label{eq:loss}
    \begin{array}{lll}
        L_{ent}(f_t; X_t) = -\mathbb{E}_{x_t \in X_t} \sum_{k=1}^{K} d_k(f_t(x_t)) \log d_k(f_t(x_t)) \\
        L_{div}(f_t; X_t) = \sum_{k=1}^{K} p_k \log p_k = -D_{KL}(p, \mathbf{1}_K) - \log k
    \end{array}
\end{equation}
}

\subsection{Experiments}\label{sec:experiments}
\subsection{Setup}
Our optimizer is AdamW~\cite{adamw} with annealing strategy for the learning rate \cite{long2017deep} adjusted during propagation with the formula $\eta_p = \frac{\eta_0}{(1+10p)^{0.75}}$, $\eta_0 = 0.001$, $p$ being the training progress linearly increasing from $0$ to $1$. Our architecture and default hyperparameters are based on \cite{zhang-etal-2021-matching} (see Table \ref{table:hyperparams}). The use of \texttt{roberta-base}~\cite{roberta} is based on existing dialectal NLP literature \cite{held2023tada,zhang-etal-2021-matching}. Each training was done on 6 NVIDIA RTX A5000. Metrics are shown in Table \ref{table:main-result}. $\text{TTDA}_{\text{gain}}$ is the performance improvement after applying SHOT, $(\text{TTDA}_{\text{gain}} = \text{TTDA} - F)$, where F is the fine-tuned model’s score and TTDA is post-adaptation.  Our experiments are fully reproducible and the code is available\footnote{\url{https://anonymous.4open.science/r/dialect-adaptation-086F/}}.

There has been extensive work in concept drift and distribution shift literature. However, methods to measure distribution shift in textual datasets have not been explored previously. To analyze distribution shift between dialects, we propose \textbf{dialectal gap}, measured as the difference between the evaluation performance on a dialect A and a set of dialects \{B\}. Our formulation of the dialectal gap is a modification over \citet{faisal-etal-2024-dialectbench}. Given that $eval(X,Y) \in [0,1]$ outputs the performance of a model trained on dialect $X$ evaluating on dialect $Y$, we can thus measure the \textbf{dialectal gap} as $eval(A,A) - eval(A,B)$. We assume that $eval(A,A) > eval(A,B)$ given that $A \neq B$. However, in rare circumstances, $eval(A,A) < eval(A,B)$ in small amount of cases, likely due to $A$ and $B$ being too close to each other. We hypothesize that there is a relationship between the dialectal gap and $\text{TTDA}_{\text{gain}}$. Hence we perform Pearson correlation analysis between these two variables. We also perform statistical comparisons to analyze the relationship between dialect via finetuning and TTDA in Section \ref{sec:experiments}.



\subsection{Dataset}

We first consider dialect $D$ of a joint distribution $p(x,y)$ where $x \in X, y \in Y$ are the input and output space, $p_S(x^s,y^s)$ and $p_T(x^t,y^t)$ are the distributions of the source and the target dialect respectively. During test-time, we are given the source data $D_S = {(x_1^s, y_1^s), ..., (x_n^s, y_n^s)}$ and the unlabeled target data $D_T = {x_1^t, ..., x_n^t}$. \cite{liang2023ttasurvey}. TTDA aims to use self-supervised learning on $D_T$ to improve the performance of the model originally trained on $D_S$ without access to $D_S$. The most foundational of TTDA techniques is SHOT~\cite{shot}, which leverages the use of self-supervised pseudo-labeling in conjunction with information maximization in order to improve the performance of the model at test time. Previous works have shown promise in using TTDA in the NLP domain \cite{banerjee-etal-2021-self, ye-etal-2022-robust, zhang-etal-2021-matching, wang-etal-2021-efficient-test, 10.1162/tacl_a_00468, laparra-etal-2021-semeval}. In addition, most have not adapted successful TTDA techniques such as SHOT~\cite{shot}, except for \citet{zhang-etal-2021-matching} in domain adaptation.

We perform our experiments on selected GLUE~\cite{wang2018glue} tasks: CoLA~\cite{cola}, RTE~\cite{wang2018glue}, SST-2~\cite{sst2} with dialectal transformation via Multi-VALUE~\cite{ziems2023multi}. Our selection of these tasks is deliberate to ensure a diverse and representative evaluation of TTDA’s effectiveness across NLU challenges even under limited computing resources. CoLA and SST-2 are single-sequence tasks with distinct metrics: Matthew's Correlation and accuracy respectively. RTE, on the other hand, is a multi-sequence task. This variety ensures different evaluation settings, input structures, and linguistic phenomena, ideal for robust evaluation of dialect adaptation. The dialects which we select are based on a subset from \citet{ziems2023multi}: SAE, IndE, NgE, and SingE.

\subsection{Results}
When comparing the performance of using SHOT versus performing finetuning on dialectal data, we find that \textbf{SHOT always improves the performance of models trained on SAE data (see Table \ref{table:main-result}). Furthermore, SHOT commonly outperforms model trained on dialectal data.} This suggests that building dialect-robust models require the incorporation of SHOT-based self-supervised techniques in order to reverse the effect of performing self-supervised learning of base embedding models on SAE data.

Using Table \ref{tab:pearson-app}, we calculate the individual dialectal gap and perform Pearson correlation analysis of these 27 observations (see Table \ref{tab:pearson-app}) achieving a p-value of 0.0305 and a correlation coefficient of 0.4169. When considering only 9 observations with the train dataset being SAE, the p-value is 0.0041 and the correlation coefficient is 0.8455. In both cases, the p-value is statistically significant (<0.05) and the correlation coefficient is positive. The scatterplot is given in Figure \ref{fig:scatterplot}.

\textbf{Based on this evidence, we observe that there is a strong positive relationship between the dialectal gap and $\text{TTDA}_{\text{gain}}$. This means that as the dialectal gap increases, the more TTDA becomes beneficial in reducing said gap.} In addition, \textbf{When fine-tuning on a dialect $A$ (except for SAE) and evaluating on dialects $A, B_1, B_2, ...$, the evaluation on dialect $A$ (the dialect the model was original fine-tuned on) has the best performance}.


\section{Conclusion}
Due to the \textbf{dialectal gap} which exhibits in previous literature, we formally propose a mathematical definition of the concept, apply SHOT~\cite{shot} to dialectal NLP, and do a comparative study of SHOT against finetuning under different dialectal settings. Our findings indicate that: 
\begin{enumerate}
    \item When the model is fine-tuned on SAE as a source dialect, SHOT always improves the performance of models when adapting to the target dialect, even compared to models fine-tuned on the target dialect.  
    \item When the model is fine-tuned on a non-SAE source dialect $A$, it tends to have the highest performance when evaluating on the same target dialect $A$.
    \item Dialectal gap has a strong positive relationship with $\text{TTDA}_{\text{gain}}$. This means that as the dialectal gap increases, the more TTDA is beneficial in reducing said gap. 
\end{enumerate}


\begin{table}[ht]
\centering
\resizebox{0.45\textwidth}{!}{%
\begin{tabular}{@{}lllcc@{}}
\toprule
Task  & Source       & Target        & Gap    & $\text{TTDA}_{\text{gain}}$ \\ \midrule
CoLA  & IndE      & NgE    & 1.260  & 3.16     \\
CoLA  & IndE      & SingE & 5.385  & 1.93     \\
CoLA  & NgE    & IndE      & 3.220  & 3.85     \\
CoLA  & NgE    & SingE & 5.785  & -0.47    \\
CoLA  & SAE         & IndE      & 16.945 & 2.73     \\
CoLA  & SAE         & NgE    & 14.965 & 2.02     \\
CoLA  & SAE         & SingE & 21.990 & 3.47     \\
CoLA  & SingE & IndE      & 0.000  & 2.73     \\
CoLA  & SingE & NgE    & -0.350 & -1.26    \\
RTE   & IndE      & NgE    & 0.800  & -1.19    \\
RTE   & IndE      & SingE & 1.200  & 1.20     \\
RTE   & NgE    & IndE      & 0.000  & -3.18    \\
RTE   & NgE    & SingE & 1.190  & -1.20    \\
RTE   & SAE         & IndE      & -0.450 & 1.19     \\
RTE   & SAE         & NgE    & -0.840 & 1.20     \\
RTE   & SAE         & SingE & -0.050 & 0.79     \\
RTE   & SingE & IndE      & 5.180  & 0.00     \\
RTE   & SingE & NgE    & 0.000  & 1.59     \\
SST-2 & IndE      & NgE    & 0.490  & 1.50     \\
SST-2 & IndE      & SingE & 3.490  & 2.13     \\
SST-2 & NgE    & IndE      & 3.610  & 1.00     \\
SST-2 & NgE    & SingE & 4.240  & 0.38     \\
SST-2 & SAE         & IndE      & 4.450  & 0.37     \\
SST-2 & SAE         & NgE    & 3.960  & 0.88     \\
SST-2 & SAE         & SingE & 5.580  & 0.25     \\
SST-2 & SingE & IndE      & 0.860  & -0.88    \\
SST-2 & SingE & NgE    & 0.860  & 0.74     \\ \bottomrule
\end{tabular}%
}
\caption{Dialectal gap and $\text{TTDA}_{\text{gain}}$. CoLA has a range of $[-1, 1]$ so we rescaled it to $[0,1]$. All gap values are in percentage. $\text{TTDA}_{\text{gain}}$ is between the TTDA and the baseline model as finetuned on $Source$ and evaluated on the $Target$.}
\label{tab:pearson-app}
\end{table}

\section*{Acknowledgments}
The production of the code used in this paper was partially reliant on generative AI code assistant. Generative AI tools were used to perform literature search and for assisting in the wording of the paper. All AI-generation has been validated by the authors.


\section*{Limitations}
Due to computational limits, we were not able to perform evaluation on all Multi-VALUE GLUE tasks. We also were not able to do experiments on STS-B due to this being a regression task which is not yet accommodated by SHOT~\cite{shot}. Whilst computational constraints preclude tasks like QQP for the time being, we believe our chosen tasks sufficiently demonstrate the core contributions of our study (see Section \ref{sec:experiments}). We used Multi-VALUE to create dialectal transformations which only produce syntactic variation and does not fully capture natural dialectal variation (see Section \ref{sec:related}).


\bibliography{ref}

\appendix

\section{Appendix}

\begin{table}[h]
\centering
\begin{tabular}{@{}ll@{}}
\toprule
Hyperparameters            &      \\\midrule
Epochs                     & 30   \\
Batch size (per device)    & 32   \\
Label smoothing            & 0.1  \\
IM loss multiplier         & 1.0  \\
Classifier loss multiplier & 0.3  \\
Entropy epislon            & 1e-5 \\ \bottomrule
\end{tabular}
\caption{Hyperparameters for finetuning and SHOT}
\label{table:hyperparams}
\end{table}


\begin{table*}[h]
\resizebox{\textwidth}{!}{%
\begin{tabular}{@{}ccccccccccccc@{}}
\toprule
                                 & \multicolumn{12}{c}{{T}}                                                                                                                                                                                                                                   \\ \cmidrule(l){2-13} 
                                 & \multicolumn{3}{c}{IndE}                                     & \multicolumn{3}{c}{NgE}                                            & \multicolumn{3}{c}{SingE}                                         & \multicolumn{3}{c}{Average}                     \\ \cmidrule(l){2-13} 
\multirow{-3}{*}{{S}} & B                             & A              & $\text{TTDA}_{\text{gain}}$      & B                                      & A              & $\text{TTDA}_{\text{gain}}$      & B                                      & A              & $\text{TTDA}_{\text{gain}}$      & B              & A              & $\text{TTDA}_{\text{gain}}$      \\ \midrule
{SAE}                     & \textbf{67.96}                & \textbf{68.93} & \textbf{0.98} & \textit{68.91}          & \textbf{69.94} & \textbf{1.03} & 65.76                                  & \textbf{66.69} & 0.92          & 67.54          & \textbf{68.52} & 0.98          \\
{IndE}                  & \textit{66.86} &                &               & 66.01                                  & 66.64          & 0.63          & 63.50                                  & 64.93          & \textbf{1.43} & 65.46          & 65.79          & \textbf{1.03} \\
{NgE}                & 66.92                         & 66.84          & -0.09         & \textit{\textbf{69.20}} &                &               & 65.46                                  & 65.11          & -0.35         & 67.19          & 65.97          & -0.22         \\
{SingE}                & 66.34                         & 66.50          & 0.16          & 68.18                                  & 68.75          & 0.57          & \textit{\textbf{68.35}} &                &               & \textbf{67.62} & 67.62          & 0.36          \\
{Average}                 & 67.02                         & 67.42          & 0.37          & 68.08                                  & 68.44          & 0.60          & 65.77                                  & 65.58          & 0.32          & 66.95          & 67.15          & 0.43          \\ \bottomrule
\end{tabular}
}
\caption{Average evaluation on CoLA, RTE, and SST-2. Bold is for best value for specific task and setting. Italic is for best performing evaluation dialect per dialect trained. S is the source dialect and T is the target dialect.}
\label{table:average_task}
\end{table*}

\begin{figure*}[h]
\centering
\includegraphics[width=\textwidth]{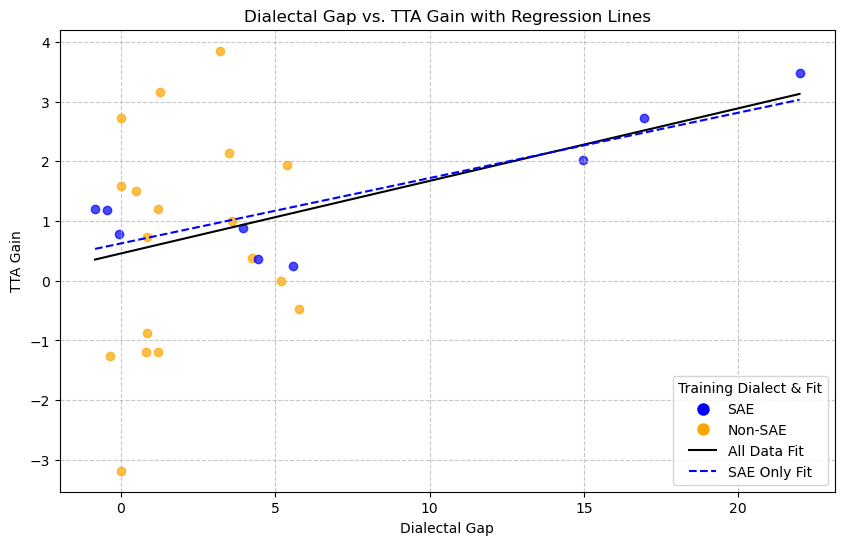}
\caption{Scatterplot of the dialectal gap against $\text{TTDA}_{\text{gain}}$. The correlation coefficient is 0.4169 and the p-value is 0.0305. The scatter plot shows positive correlation between the dialectal gap and $\text{TTDA}_{\text{gain}}$, and this is truer for SAE-only observations for reasons pointed out above. We also plot regression line for SAE-only data and all data points}
\label{fig:scatterplot}
\end{figure*}

\end{document}